\pdfoutput=1

\documentclass[11pt]{article}

\usepackage[review]{acl}

\usepackage{times}
\usepackage{latexsym}
\usepackage{graphicx} 
\usepackage{amsmath}
\usepackage[T1]{fontenc}

\usepackage[utf8]{inputenc}

\usepackage{microtype}
\newcommand\Mark[1]{\textsuperscript{#1}}

\title{Towards Multi-Turn Empathetic Dialogs with Positive Emotion Elicitation}

\author{Shihang Wang \Mark{1}\thanks{\quad The work was done when Shihang Wang was doing internship at Baidu.}, Xinchao Xu\Mark{2} , Wenquan Wu\Mark{2}, Zheng-Yu Niu\Mark{2}, \\
        {\bf Hua Wu\Mark{2} \and Haifeng Wang\Mark{2}} \\
        \Mark{1}Fu Foundation School of Engineering and Applied Science, Columbia University \\
        \Mark{2}Baidu Inc., China \\
        \texttt{sw3275@columbia.edu}\\
        \
        \texttt{\{xinchaoxu,wuwenquan01,niuzhengyu,wu\_hua,wanghaifeng\}@baidu.com}}

\begin{document}
\maketitle
\begin{abstract}
Emotional support is a crucial skill for many real-world scenarios, including caring for the elderly, mental health support, and customer service chats. This paper presents a novel task of empathetic dialog generation with positive emotion elicitation to promote users' positive emotion, similar to that of emotional support between humans. In this task, the agent conducts empathetic responses along with the target of eliciting the user's positive emotions in the  multi-turn dialog. To facilitate the study of this task, we collect a large-scale emotional dialog dataset with positive emotion elicitation, called \textbf{PosEmoDial} (about 820k dialogs, 3M utterances). In these dialogs, the agent tries to guide the user from any possible initial emotional state, e.g., sadness, to a positive emotional state. Then we present a positive-emotion-guided dialog generation model with a novel loss function design. This loss function encourages the dialog model to not only elicit positive emotions from users but also ensure smooth emotional transitions along with the whole dialog. Finally, we establish benchmark results on PosEmoDial, and we will release this dataset and related source code to facilitate future studies. 
\end{abstract}

\section{Introduction}

Emotion perception and expression are vital for building a human-like dialog system. Thanks to the availability of large-scale corpora and the rapid advances in deep learning, the potential of agents to improve the emotional well-being of users has been growing (\citealp{DBLP:journals/corr/abs-1906-09774}, \citealp{DBLP:journals/tois/HuangZG20}). In particular, the agents could provide emotional support and prevention measures in against of the increasing stress level of individuals. 

\begin{figure}[h]
    \centering
    \includegraphics[width=8cm]{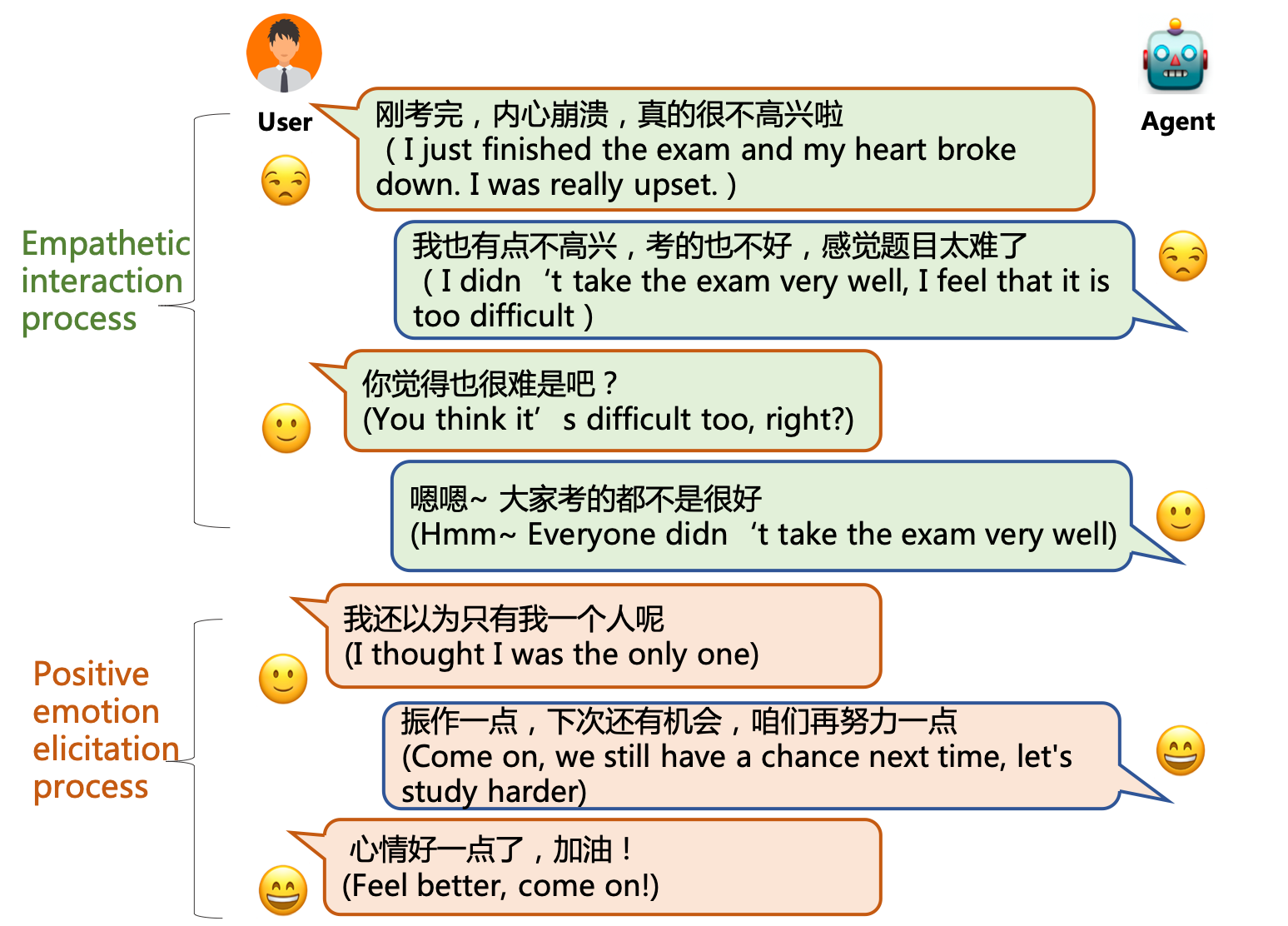}
    \caption{A sample of positive-emotion-guided empathetic conversation. It consists of two stages: (\romannumeral1) the agent expresses empathy about the situation of the user; (\romannumeral2) the agent encourages the user and changes the emotion state of the user from ``negative'' to ``positive''.}
    \label{fig:1}
\end{figure}

\begin{table*}[h]
\small
\centering
\scalebox{1.0}{
\begin{tabular}{lllllll}
\hline
\textbf{Datasets} & \textbf{\#Dialogs} & \textbf{Language}  & \textbf{Emp.} & \textbf{P.E.G} & \textbf{Multi-turn} & \textbf{Source}\\
\hline
NLPCC2017 \citep{DBLP:conf/nlpcc/HuangYZ17}& 1,119,207 & Chinese  & No & No & No & Weibo\\
MOJITALK \citep{DBLP:conf/acl/WangZ18} & 662,159 & English & No & No & No & Twitter\\
PEC \citep{DBLP:conf/emnlp/ZhongZWLM20} & 355,000 & English & Yes & No & Yes & Reddit\\
Empatheticdialog \citep{DBLP:conf/acl/RashkinSLB19} & 24,850 & English  & Yes & No & Yes & Crowd Sourcing\\
DailyDialog \citep{DBLP:conf/ijcnlp/LiSSLCN17} & 13,118 & English  & No & No & Yes & Online Websites\\
Enhanced SEMAINE \citep{DBLP:conf/aaai/LubisSYN18} & 2,349 & English  & No & Yes & No & Crowd Sourcing\\
EmotionPush \citep{DBLP:conf/globecom/HuangK18} & 8,818 & English  & No & No & Yes & Facebook Message\\
MPDD \citep{DBLP:conf/lrec/ChenHC20} & 4,142 & English  & No & No & Yes & TV-series\\
\hline
PosEmoDial (our dataset) & 819,391 & Chinese  & Yes & Yes & Yes & Web\\
\hline
\end{tabular}}
\caption{\label{EmoData-Sum}
Comparison of our dataset PosEmoDial with other datasets for emotional dialogs. Emp. denotes dialog empathy and P.E.G. denotes positive emotion guidance.
}
\end{table*}

The previous researches on empathetic dialog generation, which focuses on conducting natural empathetic responding by understanding and acknowledging any implied feelings of users sheds light on enhancing user-agent emotional bond (\citealp{DBLP:conf/acl/RashkinSLB19}, \citealp{2020Towards}). In \citealp{DBLP:conf/acl/RashkinSLB19}, a benchmark and dataset is proposed to make the dialogue system towards empathetic conversation. However, the user's emotional state at the end of the conversation are not sufficiently taken into account since current approaches only consider conducting empathetic responding in every turn of the dialogue. These models look backwards in the conversation context and might fail to jump out of user's negative emotion topics, limiting their applications in real-world scenarios, such as  providing emotional support and caring for the elderly  \citep{DBLP:conf/acl/ZhangD20}. 

Apart from that, positive emotion elicitation, which advance the conversation towards optimistic state to equip users to cope with the situation is also significantly related to positive outcomes of human interactions (\citealp{Mishara2007Which}, \citealp{2010Crisis}, \citealp{8649596}). Recently the studies (\citealp{DBLP:conf/iwsds/LubisSY017},\citealp{DBLP:conf/aaai/LubisSYN18}, \citealp{DBLP:conf/ijcai/LiFWS0W20}) drew on an important potential of  positive emotion elicitation  in maximizing user emotional experience and promoting positive emotional states, similar to that of human beings. But these works usually attempt to conduct emotion elicitation in a single turn, yielding unnatural emotional transitions and thus failing to "reach an understanding" of the individuals with the absence of backwards empathetic reflection(\citealp{Rogers2007The}, \citealp{2000Client}, \citealp{DBLP:conf/iwsds/LubisSY017}). Therefore, an ideal positive emotional elicitation process should progressively seek a certain degree of emotional resonance with the user (such as similar experiences, feelings) before improving user emotion towards a better state \citep{DBLP:conf/acl/ZhangD20}. The multi-turn empathetic dialogs with positive emotion elicitation might yield mutually reinforcing advantages for agent's empathy   and  functionality of emotional support, which is less studied in previous work.

To sum up, we present a novel task, multi-turn empathetic dialog generation with positive emotion elicitation. In this task, the agent will first conduct empathetic responding and then naturally switch to positive emotion elicitation from users. Figure~\ref{fig:1} provides an example for this task. To address this task, we encounter two challenges: (1) how to effectively capture emotions in an accurate and explainable way, (2) how to ensure smooth emotional transitions along with the whole dialog. 

To facilitate the study of this task, we collect a human-to-human multi-turn Chinese dialog dataset with positive emotion elicitation (\textbf{PosEmoDial}). In PosEmoDial, every dialog is initiated by a speaker with either a positive, neutral, or negative emotion and ends up with a positive emotion of the same speaker that is elicited by another speaker. This dataset is collected from real web users in a web forum, not being annotated by crowdsourcing, which contains more natural dialog logic about how speakers successfully fulfill positive emotion elicitation (corresponding to \emph{the second challenge}).

To address this task, we propose a novel Positive-emotion-guided empathetic dialog model (\textbf{PEGE}) by improving traditional negative log-likelihood (NLL) loss. Specifically, we introduce a new loss term, the Positive Emotion Guidance (\textbf{PEG}) loss, which measures how smoothly candidate responses at each dialog turn move from an initial emotion state at the first turn to the targeted positive emotion state at the last turn (corresponding to \emph{the second challenge}). To enable PEG loss to measure the above emotional transitions more effectively, we employ an external resource, Valence-Arousal-Dominance (\textbf{VAD}) Lexicons \citep{DBLP:conf/acl/Mohammad18}, for representation of emotions in utterances (\emph{the first challenge}). Our PEG loss encourages the dialog model to conduct positive emotion elicitation and also ensure smooth emotional transitions along with the whole dialog.

This work makes the following contributions:
\begin{itemize}
\item We present a novel task of empathetic dialog generation with positive emotion elicitation. 
\item We provide a large-scale empathetic dialog dataset with positive emotion elicitation, PosEmoDial.
\item We propose a positive-emotion-guided pre-training-empowered dialog generation model (PEGE) with novel loss function design and confirm its effectiveness.
\end{itemize}

\section{Related Work}

\textbf{Models for Emotional Dialogs} Previous work on emotional dialogs fall into three categories: (1) controlled emotional
dialog generation (\citealp{DBLP:conf/naacl/HuangZTD18}, \citealp{DBLP:conf/aaai/ZhouHZZL18}, \citealp{DBLP:conf/naacl/ColomboWMKK19}, \citealp{DBLP:conf/acl/SongZLXH19}, \citep{DBLP:conf/acl/WangZ18},\citealp{DBLP:conf/acl/ShenF20}, \citealp{DBLP:journals/corr/abs-2012-08377}); (2) empathetic dialog generation (\citealp{DBLP:conf/acl/RashkinSLB19}, \citealp{DBLP:conf/emnlp/LinMSXF19}, \citealp{DBLP:conf/emnlp/MajumderHPLGGMP20}, \citealp{2020Towards}); (3) emotion elicitation (\citealp{DBLP:conf/aaai/LubisSYN18}, \citealp{DBLP:conf/ijcai/LiFWS0W20}, \citealp{DBLP:journals/corr/abs-1906-08487}). Our model can conduct positive emotion elicitation, while previous work on empathetic dialog generation might fail to fulfill this dialog goal. Moreover, we emphasize natural emotional transitions through multi-turn dialogs, which is neglected by previous works on emotion elicitation.

\textbf{Datasets for Emotional Dialogs} To facilitate the study of emotional dialog, many researchers have created multiple datasets in previous works, as shown in Table~\ref{EmoData-Sum}. The two large-scale automatic annotated dataset NLPCC2017 \citep{DBLP:conf/aaai/ZhouHZZL18} and MOJITALK \citep{DBLP:conf/acl/WangZ18} and the manually labeled dataset DailyDialog \citep{DBLP:conf/ijcnlp/LiSSLCN17} are widely used for controlled emotional dialog generation (\citealp{DBLP:conf/aaai/ZhouHZZL18}, \citealp{DBLP:conf/acl/WangZ18}, \citealp{DBLP:journals/ai/WangW19}, \citealp{DBLP:conf/acl/ShenF20}). The Empatheticdialog \citep{DBLP:conf/acl/RashkinSLB19} dataset is designed for training empathetic dialog models (\citealp{DBLP:conf/emnlp/LinMSXF19}, \citealp{DBLP:conf/emnlp/MajumderHPLGGMP20}, \citealp{2020Towards}). The Enhanced SEMAINE dataset \citep{DBLP:conf/aaai/LubisSYN18} is constructed for the study of emotion elicitation by selecting or rewriting dialogs that can elicit positive emotion from SEMAINE corpus. In comparison with Empatheticdialog and Enhanced SEMAINE, our dataset is collected from dialogs between real web users, not through crowdsourcing. Then our dataset contains more natural emotional transitions logics with empathy and emotion elicitation naturally expressed. In addition, our dataset size is among the largest ones.

\begin{table}
\centering
\small
\scalebox{0.9}{
\begin{tabular}{lllll}
\hline
\textbf{Context Emo} & \textbf{Negative} & \textbf{Neutral} & \textbf{Positive} & \textbf{Total} \\
\hline
\emph{\#Session} & 220,136 & 403,507 & 195,748 & 819,391\\
\emph{\#Utterance} & 868,658 & 1,581,445 & 725,426 & 3,175,529 \\
\hline
\end{tabular}
}
\caption{\label{DuElicDial-Sum}
Data scale of PosEmoDial, where Context Emo address the emotion of the first utterance by speaker. All sessions in PosEmoDial have at least three utterances (before deleting the last utterance), and the last utterance by user must be optimistic.
}
\end{table}

\section{Dataset Construction}
\label{sec:3}

\subsection{Task Definition}

The person who starts the dialog is regarded as \textbf{user}, and the other one is regarded as \textbf{agent}. The goal of our task is to conduct empathetic dialog generation with positive emotion elicitation. There are two main characteristics of this task. Firstly, from the perspective of dialog goals, the agent should successfully elicit positive emotions from users through multi-turn dialogs. If the emotion state of users at the first dialog turn is negative or neutral, the agent should lead the dialog to a positive emotion state. If the initial one is positive, the agent should keep the emotion state to be positive or neutral. Secondly, from the perspective of emotional changes, the dialogue should be conducted in a natural, empathetic and gradual way. 

\subsection{Data Collection}
\label{subsec:datacollect}
In this work, we collect the dataset from natural dialogs of real web users on public websites, instead of through data annotation by crowdsourcing. The reason is that the empathy expressing of real users are more natural, and their chatting topics are more close to everyday life scenarios. We first collect Chinese dialogs from public social media and implement similar data cleaning process as done in \citet{DBLP:journals/corr/abs-2006-16779}, which yielding a dataset containing 1.2 billion two-people dialog sessions. Then we introduce an ERNIE \citep{DBLP:journals/corr/abs-1904-09223} based TextCNN \citep{DBLP:conf/emnlp/Kim14} model to recognize the emotion of each utterance in dialogs. The detailed filtering procedures on the raw dataset are shown as follows:

1) The first utterance and the last utterances are from the same speaker who plays the role of user.

2) The probability of any negative or neutral or positive emotion in the first utterance is greater than 0.5. It helps us to improve the quality of emotion polarity information that is identified on this dataset.

3) The probability of any positive emotion in the last utterance is greater than 0.9. It also helps us to improve the quality of emotion related automatically-annotated information.

4) Delete dialogs with non-emotion related topics, such as renting, job hunting, blind date, which are not related to emotion eliciting but generally end up with positive utterance like "thanks" or "good" etc. (via keywords detection).

5) Delete dialogs with specific persons, institutions, address (being recognized with the use of Name Entity Recognition tools \citep{DBLP:conf/naacl/LampleBSKD16}) for privacy consideration.

6) Delete dialogs with offensive language \citep{DBLP:conf/emnlp/Kim14} to decrease the probability of generating offensive responses.

Finally, we collect 819,391 dialogs that start with any possible negative or neutral or positive emotion and end with a positive emotion, which we called PosEmoDial. Its statistics is provided in Table~\ref{DuElicDial-Sum}.

\subsection{Data Processing}

To learn how agent-side speakers conduct successful positive emotion elicitation, we delete the last utterance (from the user-side speaker) of each dialog, and require the model to predict agent-side response at each turn.

We denote the context as $\mathnormal{\{u_1,...,u_n\}}$, the ground-truth response as ${r}$, the generated response as ${r'}$. For the sake of practicality, we treat the probability of the ${u_1}$ being emotionally positive ${p(pos|u_1)}$ or negative ${p(neg|u_1)}$ as the initial emotion state of the user-side speaker. For model training, we concatenate ${p(pos|u_1)}$ and ${p(neg|u_1)}$ with context and 
ground-truth response as the input.

\section{Our Approach}

\begin{figure*}[htp]
    \centering
    \includegraphics[width=14cm]{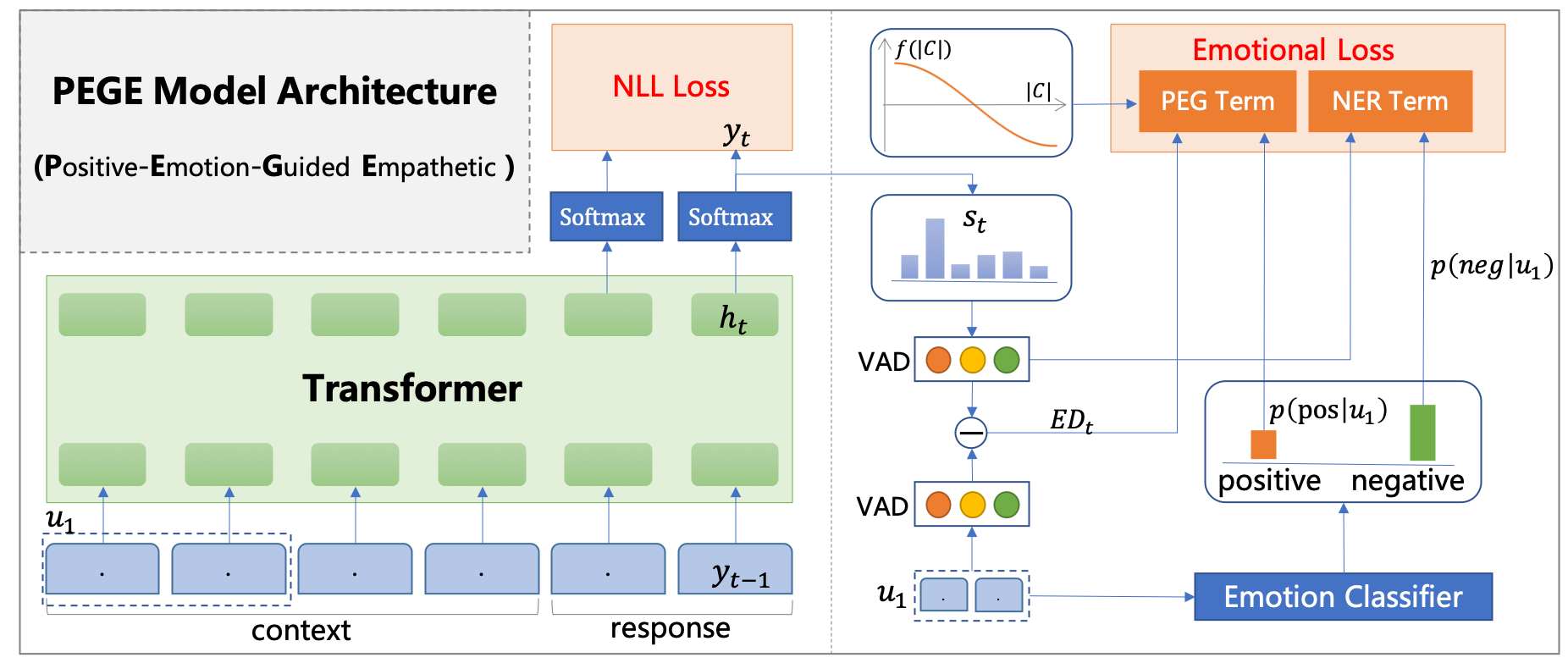}
    \caption{Illustration of our PEGE Model}
    \label{fig:2}
\end{figure*}

The proposed model is based on  PLATO-2 \citep{DBLP:journals/corr/abs-2006-16779} where we only use the General Response Generation Stage\footnote{There are two stages within the PLATO-2 model, the first stage conduct candidate responses generation and the second stage conduct responses selection. We only implement our work on the first stage of PLATO-2.} from PLATO-2 and improve its original loss function. The framework of our model is illustrated in Figure~\ref{fig:2}.  Our proposed loss function consists of two components. The first one is traditional negative log-likelihood (NLL) loss.  To effectively capture emotions in an accurate and explainable way and ensure smooth emotional transitions along with the whole dialog flow, we introduce two novel loss terms, the Positive Emotion Guidance (PEG) loss and Negative Emotion Regularization (NER) loss. The details of our model will be described in the followings.

\subsection{Emotional Distance Calculation with VAD Lexicon}
\label{sec:4-1}

Previous works have shown the effectiveness of Valence-Arousal-Dominance (VAD) Lexicons for emotional dialog generation (\citealp{DBLP:conf/aaai/Zhong0M19}, \citealp{DBLP:conf/naacl/ColomboWMKK19}, \citealp{DBLP:journals/corr/abs-2012-08377}, \citealp{2020Towards}). We further validate the high accordance between VAD score and emotion polarity obtained by a well-trained ERNIE2-TextCNN emotion classifier (\citealp{DBLP:journals/corr/abs-1904-09223}, \citealp{DBLP:conf/emnlp/Kim14}). Therefore, for the sake of token-level generation control and model efficiency, the lexicon-based VAD vectors rather than neural network-based utterance representation is selected for emotion representation in our approach. We utilize the latest and largest VAD Lexicon, the NRC\_VAD by \citet{DBLP:conf/acl/Mohammad18}, where Valence, Arousal, and Dominance are represented by continuous values in 0-1, indicating Negative to Positive, Claim to Excited, and Submissive to Dominant respectively. This lexicon includes 20,000 English vocabularies and their corresponding 13,870 distinct Chinese vocabularies. However, as there are 30k BPE tokens for the PLATO-2 lexicon. To fill this gap, we extends the NRC\_VAD to cover all the PLATO-2 lexicon.

We define \textbf{Emotional Distance (ED)} as emotional changes across different utterances. Specifically, we employ the VAD lexicon to calculate the distance between the user initial emotion state and the generated response via a 2-Norm function, as shown in Eq.(\ref{eq:wang1}).

\begin{small}
\begin{equation}\label{eq:wang1}
\begin{aligned}
ED_t=\|\sum_{j=1}^{|u_1|}\frac{\mathbf{o_{u_{1,j}}}}{|u_1|}-\sum_{i=1}^{|V|}{\mathbf{s_{t,i}o_{w_i}}}\|_2,
\end{aligned}
\end{equation}
\end{small}

the first term calculates the expected VAD value of word in $u_1$, where $|u_1|$ denotes the length of the first utterance $u_1$, $u_{1,j}$ is the j-th word in $u_1$, $\mathbf{o_{u_{1,j}}}\in \mathbf{R^3}$ is a 3-dim vector representing emotion associated with the word $u_{1,j}$ in VAD space. The second term calculate the expected VAD value of the generated word at time step $t$, where $\mathbf{s_t}=\mathbf{softmax(h_t)}$ ($\mathbf{s_t}\in \mathbf{R^{|V|}}$) is a confidence of the system of generating words $w_1,...,w_{|V|}$ at time $t$. $\mathbf{o_{w_i}}\in \mathbf{R^{3}}$ is the VAD vector of  i-th word  in the vocabulary $[ \mathbf{o_{w_1}};...;\mathbf{o_{w_{|V|}}}]$.

With the help of emotional distance, the new loss function is designed to ensure smooth emotional transitions along with the whole dialog flow as follows.

\subsection{Positive Emotion Guidance Loss}
The basic idea of the PEG loss is using emotional distance to control the emotion of response. The design process of PEG loss is described as follows: 

1) If the user's starting emotion state is positive, the emotional state of the response is expected to align with the starting emotion state to keep the positive emotion of user in the whole dialogue. The PEG loss is designed as $\sum_{t=1}^TED_t$ , which will control the emotion of the response to be close to the starting emotion state, where $ED_t$ is the measurement of emotional distance between the starting utterance and the generated response at time step t as illustrated in Eq.(\ref{eq:wang1}).

2) If the user's starting emotion state  is negative,  the response is expected to express empathy at the dialogue's initial stage, and progressively transit to positive emotional state to elicit the user's positive emotion. Therefore, the emotional distance is required to be progressively increased throughout the whole dialog. 

In order to progressively increase the emotional distance, we further improve the PEG loss by introducing a novel controlling function $f$, named as \textbf{Dialog Progress Function}.  The $f(\cdot)$ is defined as:

\begin{small}
\begin{equation}\label{eq:wang4}
\begin{aligned}
f(|C|)=\cos(\frac{\pi}{max\_turn}|C|),
\end{aligned}
\end{equation}
\end{small}

where $max\_turn$ is the maximum number of turns in dialog contexts, and it is set to 7 in our experiments. $|C|$ denotes  the number of turns in dialog context at current time step. The $f(\cdot)$ value will transit from positive value to negative value as contexts get longer as shown in the middle part of Figure 2. 

With the dialogue progress function, the PEG loss is redesigned as  $\sum_{t=1}^T[f(|C|)\cdot{ED_t}]$. Then the emotion of the response will be controlled as follows: 
\begin{itemize}
\item At the dialogue's initial stage, the emotional distance will be encouraged to be small. In other words, the emotion of response is controlled to align with the user's starting emotion to express empathy.
\item At the dialogue's latter stage, the emotional distance will be encouraged to be big because of the negative value of function $f(|C|)$ results in negative loss. In other words, the emotion of response is controlled to be different from the starting emotion of user, which will be positive.
\item At the whole dialogue stage, the emotional distance will be progressively increased from a small value to a big value because of the progressive transition of function $f(|C|)$. In other words, the emotion of response is controlled to express empathy at the dialogue's initial stage, and progressively transit to positive emotional state to elicit the user's positive emotion.
\end{itemize}

Finally, we use the probability of positive emotion of $u_1$ to combine the two kinds of the PEG loss as: 

\begin{small}
\begin{equation}\label{eq:wang3}
\begin{aligned}
L_{peg}=&\sum_{t=1}^T[p(pos|u_1)\cdot{ED_t}\\&+(1-p(pos|u_1))\cdot{f({|C|})}\cdot{ED_t}],
\end{aligned}
\end{equation}
\end{small}

if a dialog starts with a positive emotion, $p(pos|u_1)$ will be close to 1, and the first term will play a leading role. If a dialog starts with a negative emotion, $p(pos|u_1)$ will be close to 0, and the second term will play a leading role. Otherwise, both will work.

\subsection{Negative Emotion Regularization Loss}
The potential drawback of the PEG loss is that the emotion of generated responses is required to align with $u_1$ at the initial stage. Therefore, the higher the probability of negative $u_1$ is, the more likely the PEG loss will encourage the generation of negative words at the initial dialog stage. Sometimes the responses containing these words can be injurious and offensive to users. 

To address this issue, we add a NER loss to penalize the generation of too negative words with too small VAD values. The NER loss will be activated when $u_1$ is negative to balance the negative effect of the PEG loss. The NER loss is defined as: 

\begin{small}
\begin{equation}\label{eq:wang5}
\begin{aligned}
L_{ner}=\sum_{t=1}^T{p}(neg|u_1)\cdot\|\sum_{i=1}^{|V|}{\mathbf{s_{t,i}o_{w_i}}}\|_2,
\end{aligned}
\end{equation}
\end{small}
where the notation is the same as described in the above PEG loss section.

\subsection{Our Final Loss Function}

The objective of the PEGE model is to minimize the following integrated Positive-emotion-guided Empathetic Loss (PEGE Loss) $L_{pege}$:

\begin{small}
\begin{equation}\label{eq:wang6}
\begin{aligned}
L_{pege}=L_{NLL}^{Baseline}+\alpha\cdot L_{peg}-\beta\cdot L_{ner},
\end{aligned}
\end{equation}
\end{small}

where $L_{NLL}^{Baseline}$ denotes the NLL loss:

\begin{small}
\begin{equation}\label{eq:wang7}
L_{NLL}^{Baseline}=-\sum_{t=1}^T\log p(r_t|c,r'_{<t}),
\end{equation}
\end{small}

where $T$ is the length of the target response $r$ and $r'_{<t}$ denotes previously generated words before time $t$.

The hyper parameter $\alpha$ and $\beta$ in Eq.(\ref{eq:wang6}) denote the weights of PEG and NER loss respectively. We set $\alpha=5$ and $\beta=2$ for our final model based on grid search experiments.

\section{Experiments}
Following \citep{DBLP:conf/acl/RashkinSLB19} , we conduct both automatic and human evaluations for dialog systems. Human evaluation is more convincing, as automatic metrics don’t correlate well with human judgments of dialog quality \citep{DBLP:conf/emnlp/LiuLSNCP16}. 

\subsection{Evaluation Metrics}

\textbf{Automatic evaluation metrics.} Though BLEU and DISTINCT are two traditional metrics (\citealp{DBLP:conf/naacl/LiGBGD16}, \citealp{DBLP:conf/emnlp/LinMSXF19}), they have long been argued against its efficacy in open-domain dialogue generation \citep{DBLP:conf/emnlp/LiuLSNCP16}, and either BLEU or DISTINCT is less relevant to our task. We keep them mostly as a reference.

To evaluate the efficacy of our model, we define three novel metrics that we describe next to account for the positive emotion guidance capability and emotion empathy capability.

\textbf{PEG-Score:} a new metric on a scale of [0,3] to measure the positive emotion guidance capability. It rewards the positive emotion the user obtained in the last half of utterances, i.e., $U_{user}^{last}={\{u_{-2},u_{-4},...,u_{-n/2}\}}$, and calculate the adjust averaged VAD values of each word in $U_{user}^{last}$. Sum up the averaged VAD values to obtain the PEG-Score:
        
\begin{small}
\begin{equation}\label{eq:wang8}
PEG_{Score}=\sum_{VAD}\sum_{k\in U_{user}^{last}}\sum_{j=1}^{|u_k|}\frac{\mathbf{o_{u_{k,j}}-\overline{o_{vad}}}}{|{u_k}|},
\end{equation}
\end{small}
        
\textbf{E-Score:} a new metric on a scale of [-3,0] to measure the emotion empathy capability. It penalizes the emotional distance between the agent responses and the user starting utterance ($u_1$) in the first half utterances, i.e., $U_{agent}^{first}={\{u_{2},u_{4},...,u_{n/2}\}}$, and calculates the averaged VAD values of each word in $U_{agent}^{first}$. We also calculate the averaged VAD for each word in $u_1$ as the starting emotion state. Then we subtract the two values and get their absolute VAD values. Sum up the absolute VAD values to obtain the E-Score:
        
\begin{small}
\begin{equation}\label{eq:wang9}
E_{Score}=-\sum_{VAD}|\sum_{j=1}^{|u_1|}\frac{\mathbf{o_{u_{1,j}}}}{|{u_1}|}-\sum_{k\in U_{agent}^{first}}\sum_{j=1}^{|u_k|}\frac{\mathbf{o_{u_{k,j}}}}{|{u_k}|}|,
\end{equation}
\end{small}
        
\textbf{PEGE-Score:} to balance the evaluation of positive emotion guidance and empathy, we sum up PEG-Score and E-Score to obtain the PEGE-Score (on a scale of [-3,3]):
        
\begin{small}
\begin{equation}\label{eq:wang10}
PEGE_{Score}=PEG_{Score}+E_{Score},
\end{equation}
\end{small}

\textbf{Human evaluation metrics.} We run crowd-sourcing tasks at the level of both utterances and dialogs. Three crowd-sourcing workers are asked to score the response/dialog quality with a value of 0 or 1, and the final score is determined through the majority voting. These criterias are provided as follows:
 
\textbf{Coherence}: As an utterance level metric, it measures if the response is fluent, relevant and consistent with the context.

\textbf{Informativeness}: As an utterance level metric, it evaluates if the response is informative. 

\textbf{Positive emotion guidance}:  As a dialog level metric, it evaluates if the agent successfully guides the users from a non-positive emotion state to a positive emotion state, or keep their positive emotion state unchanged.

\textbf{Empathy}: As a dialog level metric, it is only measured when the positive emotion guidance score is 1 (else 0). It measures if the agent expresses empathy towards the user before positive emotion guidance, or keep the positive user not change as the criteria for positive emotion guidance.
    
\subsection{Baselines}

We select \textbf{MoEL} \citep{DBLP:conf/emnlp/LinMSXF19} and \textbf{MIME} \citep{DBLP:conf/emnlp/MajumderHPLGGMP20}, two state-of-the-art baselines which solely introduce emotion as auxiliary information like our model in empathetic dialog generation tasks. \textbf{PLATO-2} (1.6B) \citep{DBLP:journals/corr/abs-2006-16779} and \textbf{PLATO-2-FT} (fine-tuned version of PLATO-2 (1.6B) on PosEmoDial) which hold similar structure as our model are also selected. 

However, since both MoEL and MIME are trained on the English dataset Empatheticdialog \citep{DBLP:conf/acl/RashkinSLB19}, we retrain them on PosEmoDial. For the sake of comparability, the semantic word embeddings of MoEL and MIME are initialized with the PLATO-2 embeddings (2048 dimensions). 

\subsection{Results}
In multi-turn dialogue tasks, self-chat is a commonly used method to simulate human-bot conversations (\citealp{DBLP:journals/corr/abs-1909-03087}, \citealp{DBLP:conf/eacl/RollerDGJWLXOSB21}), where a model plays the role of both partners in the conversation. For the sake of our task-specificity, we employ the original PLATO-2 model to play the role of the user. Because we want to simulate actual application scenarios as much as possible, a general "user" instead of an emotionally trained one is more appropriate. Accordingly, the candidate models will play the role of agent respectively.

The way to start the interactive conversation
needs special attention. As pointed out by \citet{DBLP:conf/eacl/RollerDGJWLXOSB21} , if starting with ’Hi!’, partners tend
to greet with each other and only cover some shallow topics in the short conversation. Therefore, we construct 100 sentences as the starting utterance of different dialogues. Each sentence provides a specific context from the user's perspective, 33 of them are negative, 34 of them are neutral, and 33 of them are positive. The agent and "user" are required to perform self-chats given the context. There are 10 turns (20 utterances) in each dialog, including the input start utterance. We carry out automatic evaluation on the 100 self-chat logs and randomly select 50 conversations from 100 self-chat logs for human evaluation.

\begin{table*}[h]
\centering
\small
\begin{tabular}{lcccc}
\hline
& \multicolumn{4}{c}{\textbf{Static Eval}}\\
\cline{2-5}
\textbf{Models} & BLEU1$\uparrow$ & BLEU2$\uparrow$ & Distinct-1$\uparrow$ & Distinct-2$\uparrow$\\
\hline
MoEL & 5.901\% & 2.077\% & 6.087\% & 19.728\%\\
MIME & 6.458\% & 2.117\% & 6.709\% & 19.372\%\\
PLATO-2 & \textbf{7.204\%} & 1.966\% & 8.418\% & 34.249\%\\
PLATO-2-FT & 7.024\% & \textbf{2.131}\% & 12.937\% & 44.512\%\\
\hline
\textbf{Ours} & 6.870\% & 2.039\% & \textbf{13.266\%} & \textbf{47.249\%}\\
\hline
\end{tabular}
\begin{tabular}{ccc}
\hline
\multicolumn{3}{c}{\textbf{Interactive Eval}}\\
\hline
PEG-Score$\uparrow$& E-Score$\uparrow$ & PEGE-Score$\uparrow$\\
\hline
0.063 & -0.214 & -0.151\\
0.077 & -0.202 & -0.125\\
-0.012 & -0.189 & -0.201\\
0.090 & -0.185 & -0.095\\
\hline
\textbf{0.160} & \textbf{-0.126} & \textbf{0.034}\\
\hline
\end{tabular}
\caption{\label{Interactive-auto}
Comparison of automatic evaluation metric results under a static 5k test set and interactive self-chat dialogs among our model and baselines.}
\end{table*}

\begin{table}
\centering
\small
\scalebox{0.9}{
\begin{tabular}{lllll}
\hline
\textbf{Models} & Coh.$\uparrow$ & Inf.$\uparrow$ & P.E.G.$\uparrow$ & Emp.$\uparrow$ \\
\hline
MoEL & 0.190 & 0.904 & 0.260 & 0.260\\
MIME & 0.228 & 0.892 & 0.300 & 0.140\\
PLATO-2 & 0.934 & \textbf{0.974} & 0.320 & 0.260\\
PLATO-2-FT & 0.916 & 0.954 & 0.460 & 0.380\\
\hline
\textbf{Ours} & \textbf{0.946} & 0.962 & \textbf{0.700} & \textbf{0.620}\\
\hline
\end{tabular}}
\caption{\label{Interactive-human}
Comparison of human evaluation metric results on self-chat dialogs among our model and baselines. Coh., Inf., P.E.G. and Prog. stand for Coherence, Informativeness, Positive emotion guidance, and Empathy, respectively.
}
\end{table}

\textbf{Automatic evaluation.} Table~\ref{Interactive-auto} provides the automatic evaluation results for all the models. First, in terms of positive emotion elicitation, it shows that our model performs the best. Our model and PLATO-2-FT, which are fine-tuned on our PosEmoDial dataset, gain substantial improvements compared to PLATO-2. It indicates the effectiveness of our dataset for improving positive emotion elicitation capability. Moreover, when comparing our model with PLATO-2-FT, it can also be noted that the PEGE loss can provide an additional improvement on positive emotion guidance capability. Therefore, we conclude that our dataset and PEGE loss can work jointly to improve positive emotion guidance capability efficiently. Second, in terms of dialog empathy, our model gains the best performance as well. Our model's significant advantage over the second-best model PLATO-2-FT verifies the effectiveness of our loss design towards empathy capability. MoEL and MIME, which are not pre-trained on the large-scale corpus, are less capable of generating appropriate responses, hurting their empathetic dialog capability and resulting in a slightly worse E-Score than PLATO-2 and PLATO-2-FT. These results confirm the efficiency of our model in positive emotion elicitation while ensuring dialog empathy.

\textbf{Human evaluation.} Table~\ref{Interactive-human} provides the human evaluation results for all the models. Our model has significantly better performance on two task-specific metrics (positive emotion guidance and empathy), considerably better performance on the coherence metric, and comparable performance on the informativeness metric. By comparing our model with PLATO-2-FT, our model obtains around 52\% improvements on P.E.G. and 63\% improvements on Emp. This remarkable result demonstrates the effectiveness of our PEGE loss on positive emotion guidance and empathy capability. Our dataset PosEmoDial also shows its effectiveness in training emotional dialog model as PLATO-2-FT fine-tuned on PosEmoDial outperforms PLATO-2 with 44\% improvements on P.E.G. and 46\% improvements on Emp. By applying PEGE loss and PosEmoDial simultaneously, our model gains 119\% improvements on P.E.G. and 138\% improvements on Emp. over PLATO-2, which further verifies the mutual benefits of our PEGE loss and PosEmoDial dataset.

Moreover, the models which get better performance on human evaluation metrics P.E.G. and Emp. also get higher scores on automatic evaluation metrics, PEG-Score, E-Score, and PEGE-Score. This result indicates the reliability of our proposed automatic metrics. We also observe that 81.37\% of dialogues that successfully guide the user towards positive emotion express empathy before emotion elicitation. It verifies our proposed dialog task's rationality, i.e., expressing empathy before transit to positive emotion elicitation is crucial for building a human-like dialog system with emotion perception and expression capability.

\subsection{Ablation Study}
To evaluate the effect of the PEG loss and NER loss, we delete them respectively or simultaneously to get $L_{ner}$, $L_{peg}$ and $L_{NLL}^{Baseline}$. We also eliminate the impact of PoSEmoDial by fine-tuning PLATO-2 and our model on 1M randomly selected dataset, denote as $D_{plato}$ and $D_{pege}$. Note that when $L_{NLL}^{Baseline}$ is applied, the model is equivalent to the settings of PLATO-2-FT.

\begin{table}
\centering
\small
\scalebox{1.0}{
\begin{tabular}{llll}
\hline
\textbf{Models} & PEG-Score$\uparrow$ & E-Score$\uparrow$ & PEGE-Score$\uparrow$ \\
\hline
$L_{NLL}^{Baseline}$ & 0.090 & -0.185 & -0.095\\
$L_{ner}$ & 0.068 & -0.177 & -0.109\\
$L_{peg}$ & 0.065 & -0.134 & -0.069\\
\hline
$D_{plato}$ & -0.011 & -0.191 & -0.202\\
$D_{pege}$ & 0.072 & -0.139 & -0.063\\
\hline
\textbf{Ours} & \textbf{0.160} & \textbf{-0.126} & \textbf{0.034}\\
\hline
\end{tabular}}
\caption{\label{Ablation-inter}
Comparison of automatic evaluation metric results under interactive self-chat dialogues among our model, ablation models, and models on random dataset. 
}
\end{table}

Table~\ref{Ablation-inter} illustrates the results of the ablation study. Our model with PEGE loss gets the best performance, and the model with $L_{ner}$ gets bad performance on all metrics. With only NER loss ($L_{ner}$) remains, the model is more inclined to generate positive responses directly instead of conditioned on the user emotion state transition, which may not necessarily lead to positive feedback from users. This result is consistent with our real-world observations that the response to a negative statement with positive emotion directly without any emotional transition sometimes is inappropriate and even offensive. As the PEG loss $L_{peg}$ is designed with both positive emotion elicitation capability and empathy capability, $L_{peg}$ performs much better. However, without NER loss, the model with $L_{peg}$ will endure the risk of generating excessively negative responses, which may sometimes be unacceptable to users as well, and therefore bring no gain with positive emotion elicitation. The results suggest that all components in PEGE loss $L_{pege}$ are valuable and indispensable.

The comparison between $D_{plato}$ and $D_{pege}$ illustrates that our model is not data-dependent and can be generalized in other datasets since considerable improvements can be obtained on all three metrics even PEGE model is trained on randomly selected data. Meanwhile, PosEmoDial can actually facilitate model performance for both PLATO-2 and PEGE, validating its effectiveness in our task. 

\section{Conclusion}
In this paper, we propose a novel task of multi-turn empathetic dialogs with positive emotion elicitation and collect a human-to-human Chinese multi-turn emotional dialog dataset with positive emotion elicitation (PosEmoDial). Then we propose a novel positive-emotion-guided empathetic dialog model (PEGE) by improving traditional NLL loss. The updated loss can encourage the dialog model to not only elicit positive emotions from users, but also ensure smooth emotional transitions along with the whole dialog flow. The results of the experiments confirm the usability of our dataset and the effectiveness of our model. In the future, we will introduce psychology-related domain knowledge to facilitate the modeling of in-depth emotional dialogs to support emotional counseling.

\section{Ethical Considerations}
We are sure that PosEmoDial has been collected in a manner that is consistent with the terms of use of any sources and the intellectual property and privacy rights of the original authors of the texts. Meanwhile, our project is approved by an IRB. Finally, we also provide details on the characteristics of PosEmoDial and steps taken to ensure the potential problems with the quality of the dataset do not create additional risks in Section ~\ref{sec:3}.

\bibliography{anthology,custom}

\appendix


\end{document}